\pdfoutput=1

\documentclass[11pt]{article}

\usepackage[]{acl}

\usepackage{svg}

\usepackage{times}
\usepackage{latexsym}
\usepackage{graphicx}
\usepackage{subcaption}
\usepackage{booktabs}
\usepackage{makecell}  
\usepackage{multirow}
\usepackage{xcolor}
\usepackage[T1]{fontenc}

\usepackage[utf8]{inputenc}

\usepackage{microtype}

\newif\ifcomments
\ifcomments
    \providecommand\uri[1]{[\textcolor{blue}{Uri: {#1}}]}
    \newcommand\matan[1]{[\textcolor{orange}{Matan: {#1}}]}
    \newcommand\roee[1]{[\textcolor{red}{Roee}]}
    \newcommand\jonathan[1]{[\textcolor{magenta}{Jonathan: {#1}}]}
    \newcommand\reut[1]{[\textcolor{teal}{Reut: {#1}}]}

    \def \ifempty#1{\def\temp{#1} \ifx\temp\empty }

\else
    \providecommand{\uri}[1]{}
        \providecommand{\matan}[1]{}
            \providecommand{\jonathan}[1]{}
                \providecommand{\roee}[1]{}
                    \providecommand{\reut}[1]{}
\fi

\title{Multilingual Instruction Tuning With Just a Pinch of Multilinguality}

\author{Uri Shaham$^{\tau\gamma}$ \quad Jonathan Herzig$^\gamma$ \quad Roee Aharoni$^\gamma$ \\ \quad \textbf{Idan Szpektor}$^\gamma$ \quad \textbf{Reut Tsarfaty}$^\gamma$ \quad \textbf{Matan Eyal}$^\gamma$ \\
\\
$^\tau$ Tel Aviv University\\
$^\gamma$ Google Research\\
}

\begin{document}
\maketitle

\begin{abstract}
As instruction-tuned large language models (LLMs) gain global adoption, their ability to follow instructions in multiple languages becomes increasingly crucial. In this work, we investigate how multilinguality during instruction tuning of a multilingual LLM affects instruction-following across languages from the pre-training corpus. We first show that many languages transfer some instruction-following capabilities to other languages from even monolingual tuning. Furthermore, we find that only 40 multilingual examples integrated in an English tuning set substantially improve multilingual instruction-following, both in seen and unseen languages during tuning. In general, we observe that models tuned on multilingual mixtures exhibit comparable or superior performance in multiple languages compared to monolingually tuned models, despite training on 10x fewer examples in those languages. Finally, we find that diversifying the instruction tuning set with even just 2-4 languages significantly improves cross-lingual generalization. Our results suggest that building massively multilingual instruction-tuned models can be done with only a very small set of multilingual instruction-responses.

\end{abstract}

\section{Introduction}

Instruction tuning is a fundamental aspect of building modern general-purpose large language models (LLMs), involving fine-tuning a pre-trained model on pairs of instructions and corresponding responses \citep{mishra-etal-2022-cross,wei2022finetuned,sanh2022multitask,NEURIPS2022_b1efde53}. For these models to be globally applicable, they must operate on a wide range of languages, yet, most instruction tuning datasets are typically limited to English.
While curating naturally occurring instructions and responses for every language is challenging, cross-lingual transfer has emerged as a promising approach, in which a model is fine-tuned using one language, and acquiring similar abilities in another
\citep{pires-etal-2019-multilingual,wu-dredze-2019-beto,artetxe-schwenk-2019-massively,K2020Cross-Lingual,conneau-etal-2020-unsupervised,conneau-etal-2020-emerging}. The ability to follow instructions for languages seen only at pre-training can significantly expand the applicability of LLMs, allowing them to be used by more people worldwide. In this work, we show that instruction-tuning of multilingual LLMs transfers across languages better  than previously known, and that even minimal language diversity in the tuning set can further unlock instruction-following generalization 
to languages that are unseen during instruction tuning.

We investigate the effect of multilingual data on instruction-following across languages using an LLM pre-trained on hundreds of languages \cite{anil2023palm}, and high-quality, open-ended instructions and responses \cite{zhou2023lima,k2023openassistant} translated into 11 languages, across different families and writing systems. Initially, we examine the transferability of monolingual instruction tuning across different languages. Naturally, tuning using each language individually enhances performance within that language. Notably, we find that this also translates into instruction-following capabilities across other languages, and that tuning with English, Italian, or Spanish yields the best average multilingual performance.

Inspired by this result, we turn to ask how much multilingual data is required to improve multilingual instruction-following, while preserving English performance. We find that replacing even just 40 English training examples with multilingual examples, significantly improves instruction-following in those languages. Surprisingly, this small amount of language-diverse examples also improves performance for languages that are only seen during pre-training and are not represented in the instruction tuning set at all.

The next question we tackle is whether increasing the number of languages in the tuning set can enhance generalization to new languages from the pre-training corpus. We find that tuning using a few languages enables better performance for languages unseen during tuning, compared to monolingual tuning with the same number of examples. 

Finally, we test two potential factors that might influence the degree of cross-lingual transfer: language similarity and the amount of language-specific pre-training data, but find no significant correlations. Overall, our results provide recipes for multilingual instruction tuning that improves cross-lingual generalization, while preserving performance on English, under a fixed budget. In particular, we find that capable multilingual instruction-following models can be tuned even with a minimal amount of multilingual data.

\begin{figure*}[t!]
\centering
  \includegraphics[width=0.99
  \linewidth]{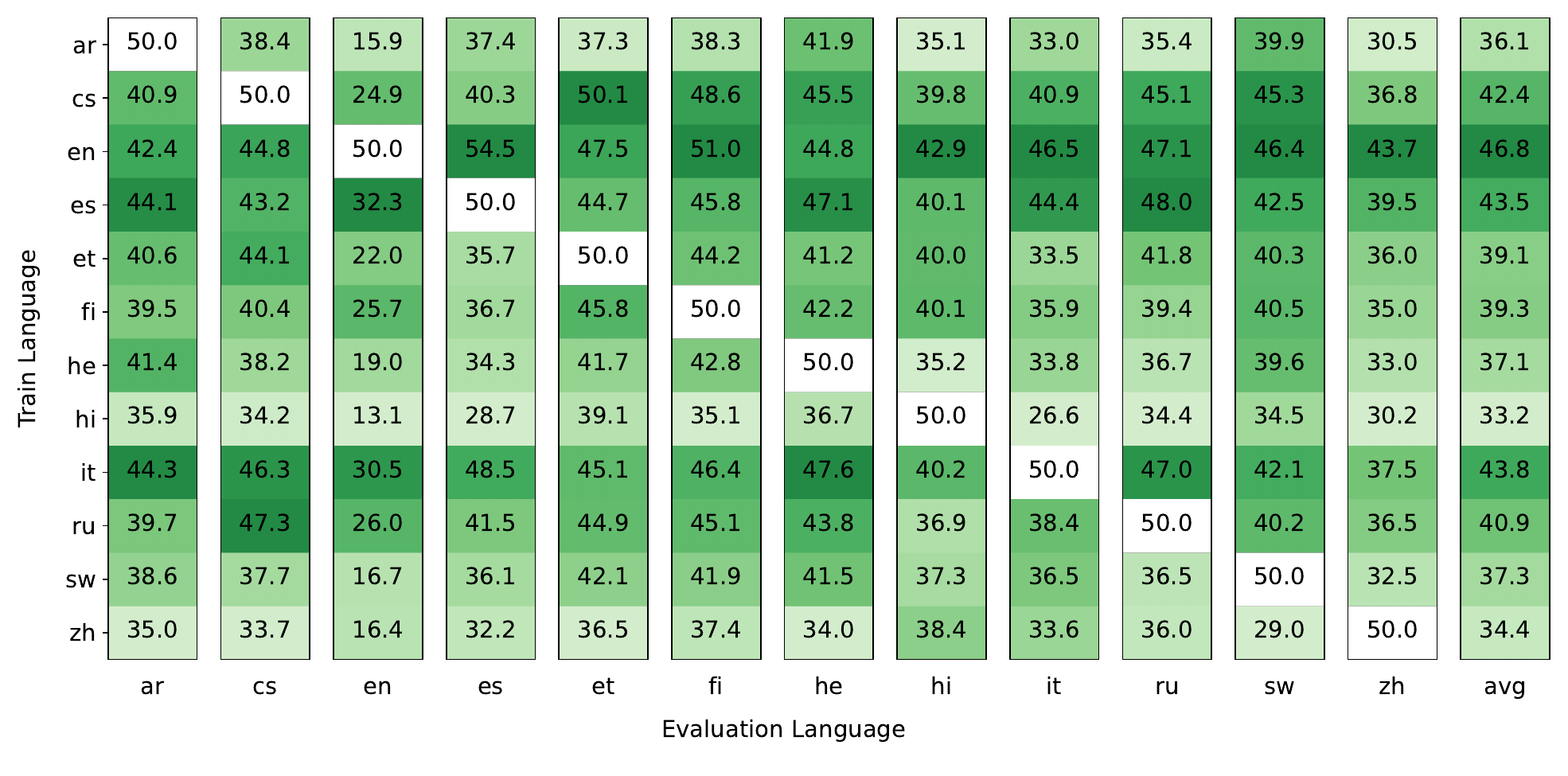}
  \caption{Per language instruction-following scores of models instruction-tuned on monolingual data. Each row
represents a model tuned using a different language, and each column is an individual heatmap of the scores of
all models on the same evaluation language. Scores are the discounted-ties weighted average of the side-by-side
scores against the model tuned on the evaluation language. The scores along the diagonal are 50 as they are the result of comparing generations to themselves, and are excluded from the heatmap
coloring.}
\label{fig:heatmaps} 
\end{figure*}

\section{Measuring Multilingual Instruction-Following}
\label{sec:measure}
Our objective is to discover how multilinguality during instruction tuning affects general-purpose instruction-following across languages. We break this down to multiple questions, including how well can monolingual instruction tuning transfer to other languages, how many multilingual examples can enhance multilingual instruction-following while preserving English performance, and whether increasing the number of languages can result in improved cross-lingual generalization. In this section we elaborate on the data, evaluation protocol, models we use, and the human annotation process to ensure the models quality.

\paragraph{Data} 
We use datasets of high-quality open-ended instructions and responses, rather than classic task-specific datasets. Our training data contains 1,000  English instructions and responses from LIMA \citep{zhou2023lima} and 3,640 from OpenAssistant\footnote{We focus on single-instruction/single-response interactions so we keep only the first prompt and response from conversations in OpenAssistant similarly to \citet{li2023selfalignment}.} \citep{k2023openassistant}. These examples resemble real world scenarios of users interacting with chatbots, with queries like "Can you explain Fermat's Last Theorem?" and "How to keep a dog hydrated?", that enable efficient tuning even with a small training set \citep{zhou2023lima}. For evaluation, we use 617 instructions from AlpacaFarm \cite{dubois2023alpacafarm}, originated from Self-Instruct \citep{wang-etal-2023-self-instruct}, Vicuna  \citep{vicuna2023}, Koala  \citep{koala_blogpost_2023}, and hh-rlhf  \citep{bai2022training}.\footnote{We exclude AlpacaFarm's evaluation instructions from OpenAssistant, as we tune using its training set.}

We use the Google Translate API\footnote{https://cloud.google.com/translate/docs/reference/api-overview} to translate the instruction-response pairs of the training set and the instructions of the evaluation set to 11 languages, creating parallel training and evaluation sets in Arabic, Chinese, Czech, English, Estonian, Finnish, Hebrew, Hindi, Italian, Russian, Spanish, and Swahili.\footnote{Languages are selected from Table 21 in \citet{anil2023palm}, describing the top-50 languages the model (\S\ref{sec:model}) was pre-trained on.}
While translated data is different from naturally sourced data per language, it allows for more control as the data size and semantics are similar for all languages. A overview of the languages, their language codes, families and scripts is described in Table~\ref{tab:lang_codes} in Appendix~\ref{sec:langs}.

\paragraph{Evaluation}
\label{sec:eval}
We conduct a side-by-side automatic evaluation protocol \citep{bubeck2023sparks,dubois2023alpacafarm,dettmers2023qlora,gudibande2023false,zheng2023judging}, in which an LLM assesses two responses for the same instruction, with the goal of identifying the superior one. We follow the common practice of presenting both responses to the model twice, alternating the order of the two responses  \citep{zheng2023judging,zhang2023plug}. The exact prompt we use is shown in Figure~\ref{fig:prompt} in Appendix~\ref{sec:sbs_prompt}. We define a ``win" for a certain response if the judge selects it twice irrespective of the order, and a ``tie" if the model selects a different response for each order. We use a discounted-tie \citep{zhou2023lima} scoring method, in which a model receives a score of 1 for a win, 0.5 for a tie, and 0 for a loss. We average the scores of individual instructions to get the score over the evaluation set and present it in percentages. To validate that the LLM judge decisions align with human preferences across languages, we conduct a human annotation study and find good aggregated agreement scores of 79.5\% for English, 77\% for Spanish, and 76.5\%, and 75\% for Russian and Hebrew, receptively. Further details on validating the LLM judge are provided in Appendix~\ref{sec:agreement}.

\paragraph{Instruction-Following Score Per Language} 
Throughout this work we measure instruction-following per language by comparing the performance of a model that was tuned on some training set $D$, to a model that was monolingually tuned on the target language $\mathcal{L}$, by using the full training set in this language, $D_{\mathcal{L} }$. Formally, we define our instruction-following ($IF$) metric for language $\mathcal{L}$:
$$
IF_{\mathcal{L}}(M_D) = S{\times}S(M_{D_{\mathcal{L}}}, M_D)
$$
Where $S{\times}S(\cdot{}, \cdot{})$ is the side-by-side protocol applied on $M_{D_{\mathcal{L}}}$ and $M_{D}$, which are the models instruction-tuned on $D_{\mathcal{L}}$ and $D$, respectively. A score of 0\% means that $M_{D}$ loses on all $\mathcal{L}$ instructions, and 50\% means the performance of $M_{D}$ and $M_{D_{\mathcal{L}}}$ in $\mathcal{L}$ are indistinguishable when aggregated over the evaluation set.

\begin{figure}[t!]
\centering
   \includegraphics[width=0.99
   \linewidth]{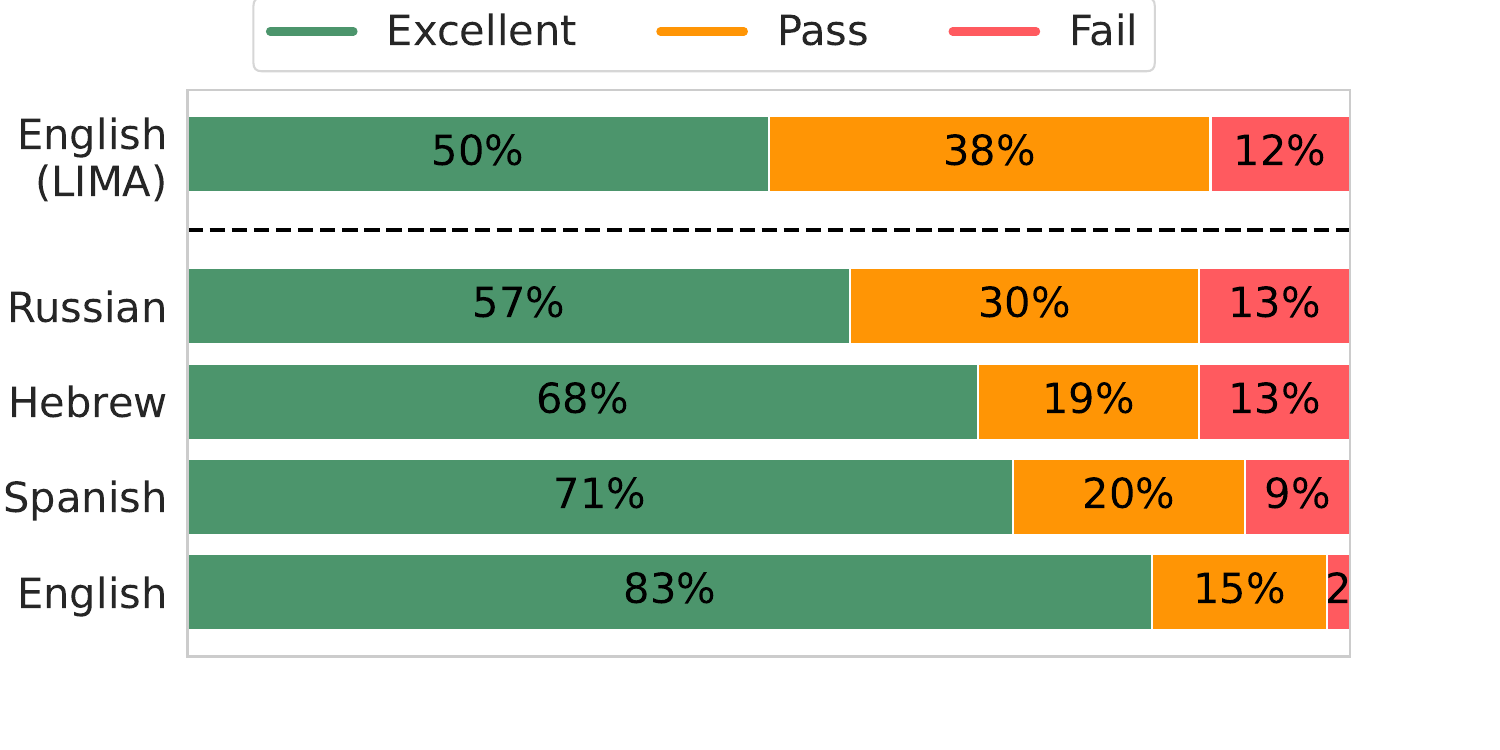}
  \caption{Human annotators rating distributions of models responses across languages. Each row describes evaluation in its corresponding language of the model tuned monolingually using that language. Numbers in the first row are reported by \citet{zhou2023lima}.}
\label{fig:abs_scores} 
\end{figure}

\paragraph{Model}
\label{sec:model}
We use the PaLM~2 model family of Transformer-based \citep{NIPS2017_3f5ee243} LLMs that were pre-trained on hundreds of languages \citep{anil2023palm}. We use PaLM~2-S as our pre-trained model for all the instruction tuning experiments, and an instruction-tuned PaLM~2-L as the judge for the side-by-side evaluation. The training and inference hyperparameters we use are described in Appendix~\ref{sec:hyperparams}. 

\begin{figure*}[t!]
\centering
  \includegraphics[width=0.99
  \linewidth]{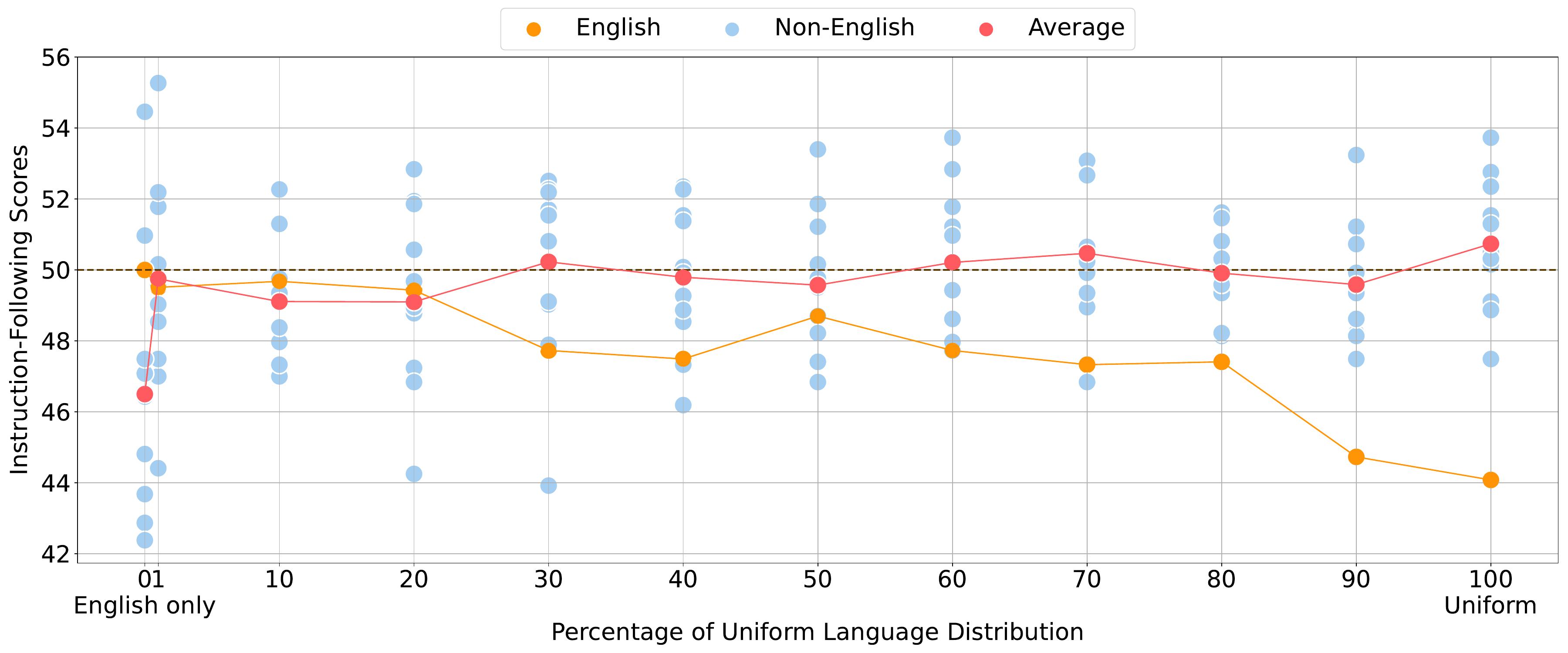}
  \caption{Instruction-following scores of models trained using  when $P\%$ of the training set is distributed uniformly across 12 languages and an $(100-P)\%$ is English only. Each X axis tick represents a tuning mixture, scores over individual non-English languages are in \textcolor[HTML]{A3CEF1}{blue}, and their averages are in \textcolor[HTML]{ff5a5f}{red}. English scores are in \textcolor[HTML]{ff9505}{orange}.}
\label{fig:en_rest12} 
\end{figure*}

\paragraph{Human Validation}
Our evaluation protocol relies on the quality of our monolingually tuned models. To validate their usage as high bar baselines in their respective languages, we conduct a human annotation study in 4 languages: English, Spanish, Russian and Hebrew. Namely, we sample 50 random instructions per language, and ask 2 native speakers to assign a score of excellent, pass, or fail \citep{zhou2023lima} to the responses generated by the model that was monolingually tuned using that language. Results in Figure~\ref{fig:abs_scores} show that our tuned models indeed demonstrate strong instruction-following abilities. Notably, the scores across languages are similar or better than the reported numbers by \citet{zhou2023lima} in English.\footnote{The differences can be attributed both to the pre-trained model and to the size of the instruction tuning dataset.} 

\section{How Much Multilinguality Is Needed For Multilingual Instruction Tuning?}
\label{sec:how_much}
We now describe our controlled experiments, designed to quantify the effect of multilingual data during instruction tuning of multilingual LLMs, following the research questions defined in \S\ref{sec:measure}. 

\subsection{Monolingual Instruction Tuning Yields Multilingual Abilities}
\label{sec:mono}

To explore zero-shot cross-lingual transfer of instruction tuning in multilingual LLMs, we tune models on a single language and evaluate them on all of the rest. We find that all of those models are able to transfer non-negligible instruction-following abilities to other languages.

\paragraph{Setup} We instruction-tune 12 models, each one using the full train set in a different language. We generate responses using every such model to the evaluation instructions in all other languages. Finally, we calculate their per language scores as described in~\S\ref{sec:eval}.

\paragraph{Results} Figure~\ref{fig:heatmaps} shows the results, where rows represent training languages and every column is an independent heatmap of the results over a single evaluation language. Most importantly, tuning using each single language yields a model with some multilingual instruction-following capabilities across languages. For context, even the model with the lowest average score, the one tuned on Hindi, achieves a score of over 30\% in 9 out of 11 cases.\footnote{For example, a score of 30\% can be obtained by wining 30\% of the instructions and losing 70\%, or by achieving a tie on 60\% of the instructions and losing 40\%.}  The model with the best average score is the one tuned on English, when Italian and Spanish also enable consistently high scores. 

Notably, we manually inspect the generations and find that our tuned models consistently respond in the same language as their instruction, regardless of the language they were instruction-tuned on, in contrast with findings in previous work \cite{touvron2023llama,chen2023monolingual}. We hypothesize that this comes from the multilingual nature of PaLM~2s' pre-training, compared to the more English-centric LLaMA  \cite{touvron2023llama}, further details are in Appendix~\ref{sec:response_lang}. In addition to our main setup, we also compare the generations of these models to the ones of the pre-trained model that was not instruction-tuned. Results shown in Figure~\ref{fig:heatmaps_base} in Appendix~\ref{sec:base_model}  further demonstrate that instruction tuning in every language separately, greatly improves instruction-following abilities across different languages.

\begin{figure*}[t!]
\centering
  \includegraphics[width=0.99
  \linewidth]{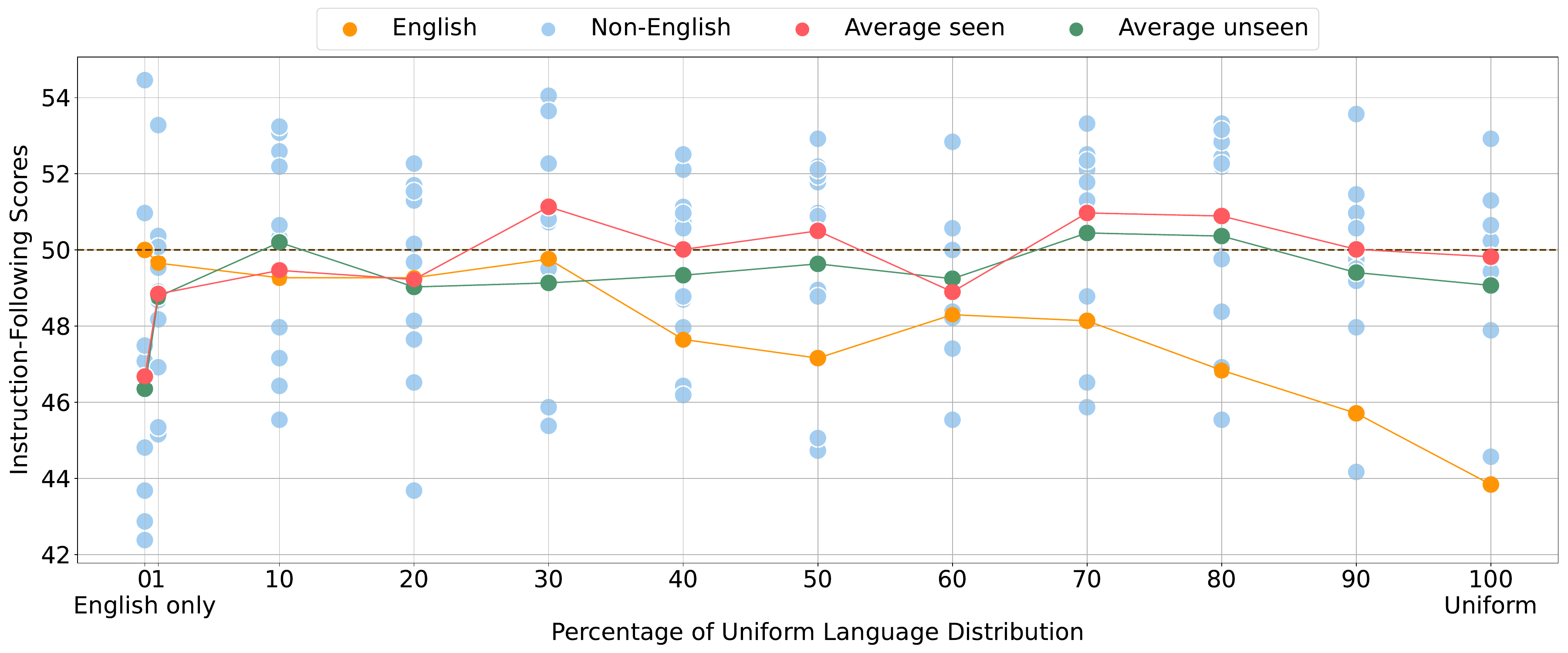}
  \caption{Instruction-following scores of models tuned when $P\%$ of the training set is distributed uniformly across 6 languages and an $(100-P)\%$ is English only. Each X axis tick represents such a tuning set, scores over individual non-English languages are in \textcolor[HTML]{A3CEF1}{blue} and English scores are in \textcolor[HTML]{ff9505}{orange}. Average scores of the 5 non-English languages in the tuning set are in \textcolor[HTML]{ff5a5f}{red}, and the average scores of the 6 languages not seen during tuning are in  \textcolor[HTML]{4C956C}{green}.}
\label{fig:en_rest6} 
\end{figure*}

\subsection{A Few Dozen Examples Improve Multilingual Instruction-following}
\label{sec:n_examples_per_lang}
Naturally, multilingual tuning, as opposed to English-exclusive tuning under a fixed training examples budget, should result in better downstream performance for non-English languages, and might hurt performance on English. Therefore, we ask how many multilingual examples can improve the instruction-following abilities across languages, while preserving English performance. To that end, we tune models on subsets of the English examples combined with subsets of multilingual examples in different ratios. We find a significant boost in multilingual instruction-following abilities even when using just a few dozen multilingual examples.

\paragraph{Setup} We create data mixtures with $P\%$ examples that are evenly split among all 12 languages, and the rest $(100-P)\%$ English examples.\footnote{Every example appears exactly once in every mixture, in a single language.} We create such a train set for every $P$ from 10 to 100, incremented by tens, and also for $P=1$, for which only 40 multilingual examples are included from across all 11 non-English languages, and the rest are English examples. Finally, we evaluate every tuned model on every one of the 12 languages as defined in~\S\ref{sec:eval}.

\paragraph{Results} Figure~\ref{fig:en_rest12} visualizes the results. As expected, multilingual examples in the train set improve the score on their languages (Red), and diluting the number of English examples hurts the performance in English (Green). Notably, the significant multilingual improvement comes from replacing only $1\%$ of the English examples by multilingual ones, which translates to 40 examples evenly distributed across the training languages. These results on the effect of such a small amount of language-diversity extend findings regarding task-diversity by \citet{zhou2023lima}, which demonstrated that a capable monolingual instruction-following model can be tuned using only 1,000 high-quality examples. A second trend is that these models often outperform their monolingually-tuned counterparts on the very language the latter were exclusively tuned on (blue markers above the 50 line). For example, the model tuned using the uniform set ($P=100$) preforms similarly or better than the individual monolingually-tuned models in 8 of 12 languages, despite being trained on 12 times less instruction-response pairs for each language. This suggests that for some languages, multilingual tuning can enable better instruction-following abilities compared to a traditional monolingual tuning with the same number of examples.

\subsection{A Few Dozen Examples Improve Cross-lingual Generalization}
\label{sec:n_examples_per_lang_2}
Combining the lessons on cross-lingual generalization from monolingual tuning and the effect of a small amount of multilingual examples from previous sections, we turn to examine how multilingual examples in the tuning set affect language generalization. 
Specifically, we conduct a similar experiment to the one in~\S\ref{sec:n_examples_per_lang}, this time using only half of the languages for tuning while the rest of languages are unseen. In line with the results from~\S\ref{sec:n_examples_per_lang}, we find that a very small amount of multilingual examples also improve performance on languages that were not in the tuning set.

\paragraph{Setup}
We repeat the setup from~\S\ref{sec:n_examples_per_lang}, this time with only English and 5 more languages: Arabic, Finnish, Italian, Russian, and Swahili, and evaluate models again on all 12 languages.

\paragraph{Results}
Results in Figure~\ref{fig:en_rest6} show similar trends to the ones in Figure~\ref{fig:en_rest12}. Specifically, the average score over non-English training languages (red) again improves very quickly, even with $P=1$. Strikingly, this is also true for languages that the model has only seen during pre-training, and are not represented at all in the instruction tuning dataset (orange). This suggests that very few multilingual examples can not only improve performance for the languages of those examples, but also enable better cross-lingual instruction-following generalization.
\subsection{Even a Small Number of Languages Improves Cross-Lingual Generalization}
 \label{sec:add_langs}
Given the results on the impact of a small number of multilingual \textit{examples} from a fixed set of languages, we ask whether a small number of \textit{languages} can also enhance cross-lingual generalization. We experiment with different numbers of languages in the tuning set and indeed observe that the transfer to languages only seen during pre-training improves from the very first additional languages.

\begin{figure}[t!]
\centering
   \includegraphics[width=0.99
   \linewidth]{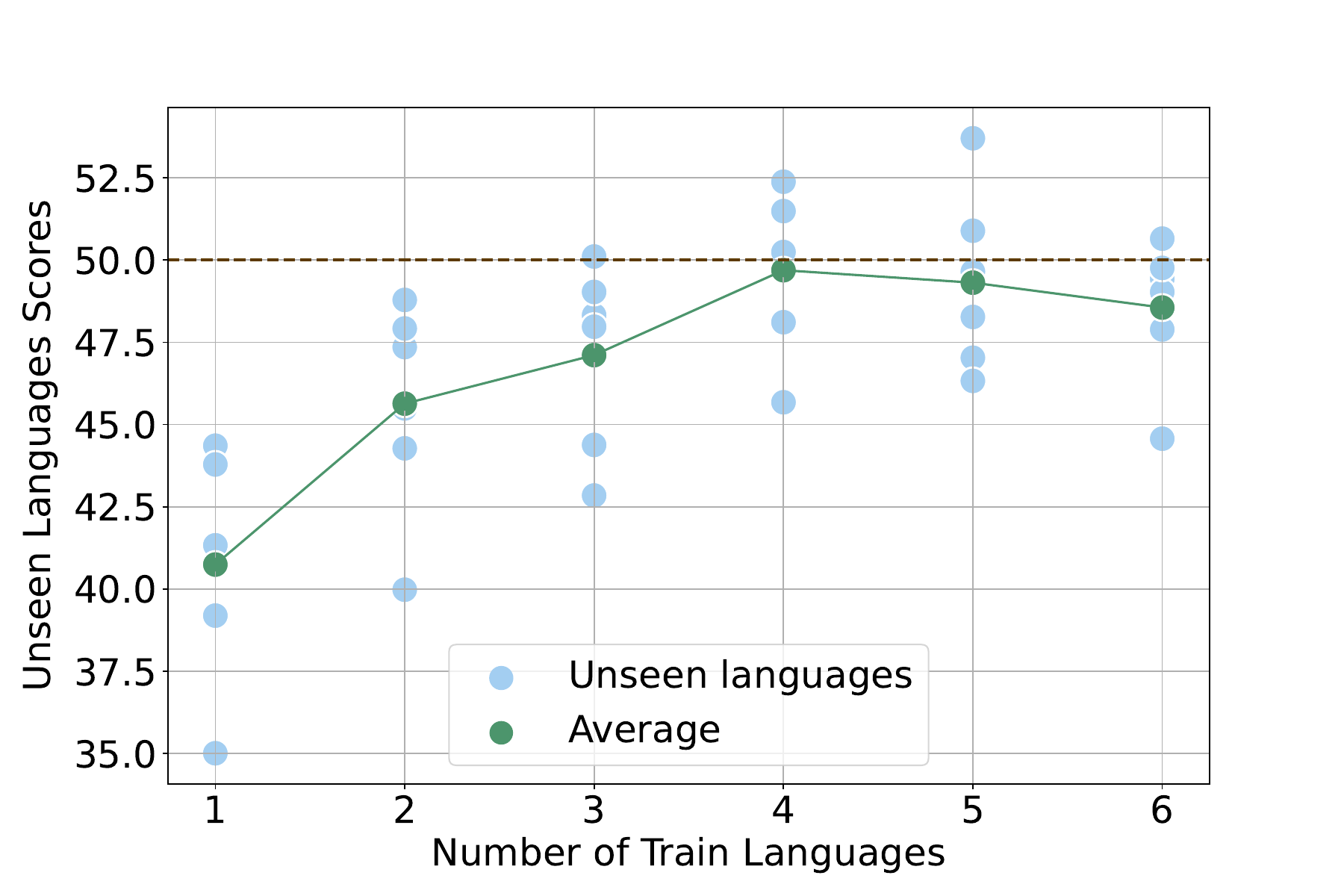}
  \caption{Instruction-following scores in Czech, Estonian, Hebrew, Hindi, Spanish, and Chinese of models instruction-tuned using various subsets of Arabic, English, Finnish, Italian, Russian, and Swahili. \textcolor[HTML]{A3CEF1}{Blue} markers are the average scores per evaluation languages across models tuned with the same number of languages. The averages of those individual languages scores are in \textcolor[HTML]{4C956C}{green}.}
\label{fig:adding_langs} 
\end{figure}

\begin{figure}[t!]
\centering
   \includegraphics[width=0.99
   \linewidth]{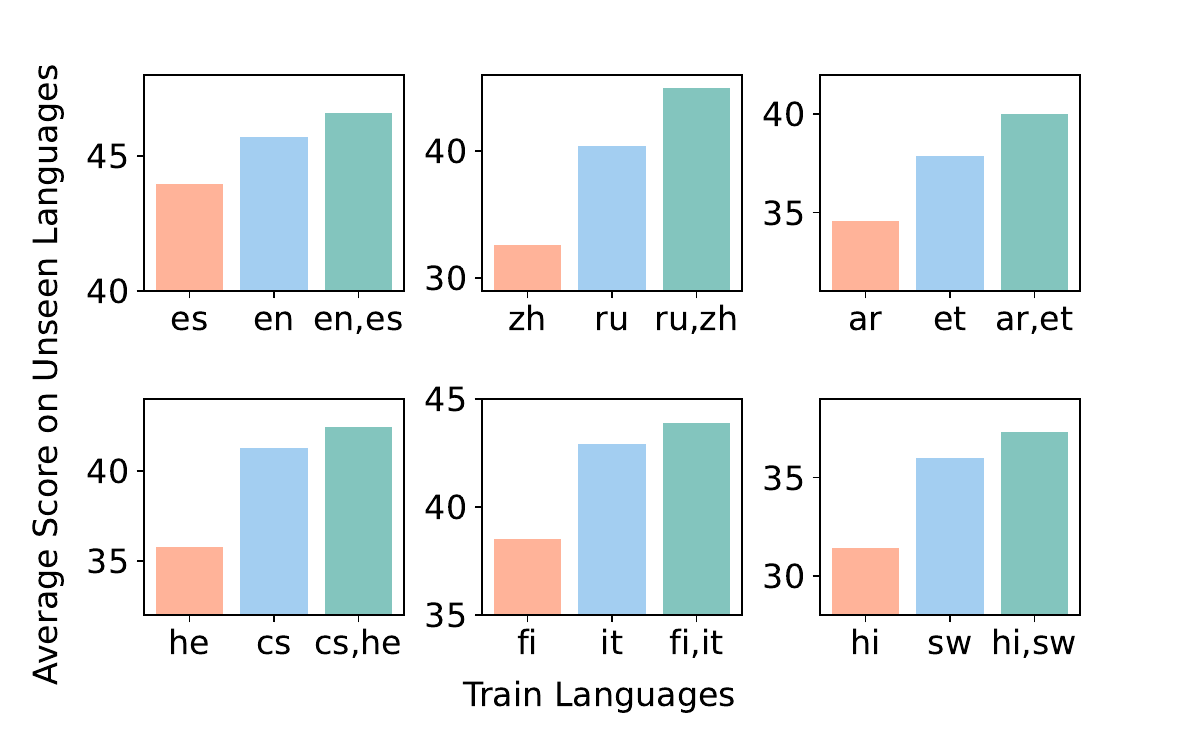}
  \caption{Average instruction-following scores of languages not seen during instruction tuning. For example, the top-left corner describes the scores of 3 models instruction-tuned on $100\%$ Spanish, $100\%$ English, and $50\%$ Spanish and $50\%$ English. The Y axis of this sub-figure is the average score across all language excluding Spanish and English.}
\label{fig:bilingual} 
\end{figure}

\paragraph{Setup}
We instruction-tune models on a single language and up to 6 languages. At each step, we add a language to the tuning set, and split the same examples budget uniformly among the current set of languages. We use the 6 training languages from \S\ref{sec:n_examples_per_lang_2}, and follow 3 different permutations that determine the order in which we add languages to the mix. These permutations are shown in Table~\ref{tab:add_langs} in Appendix~\ref{sec:lang_perms}. We evaluate every model on each of the remaining 6 languages, and average scores per evaluation language across models that are tuned using the same number of languages.

\paragraph{Results} Results on Figure~\ref{fig:adding_langs} show that adding languages to the tuning set improves cross-lingual generalization. The average score (red) increases from tuning on monolingual data to tuning on bilingual data, and even more when using 3 and 4 languages, where the average score gets to almost 50. At that point, there is an indication for saturation, as more languages does not seem to improve transfer further. These findings demonstrate that diversifying the instruction tuning data with only a few different languages can improve cross-lingual transfer to new languages, only seen during pre-training. 

\paragraph{Bilingual Tuning Sets}
To show this holds for even more combinations of languages, we randomly split all languages to pairs, and tune models using $50\%$ of the examples in the one language and $50\%$ in the other. We evaluate each of these models on the remaining 10 languages, and compare their score to the ones of the two models tuned using the full monolingual sets. Results on Figure~\ref{fig:bilingual} reveal that bilingual tuning helps generalize to new languages better than monolingual tuning.

\section{Potential Factors of Transferability}
Following the results from the previous sections, a natural question arises: what factors can predict the degree of cross-lingual transfer? We explore two immediate candidates. Initially, we examine the relation of various aspects of language similarity to transferability within language pairs. Next, we look into whether the proportion of language-specific data in the pre-training corpus correlates with the amount of cross-lingual transfer of instruction tuning using the given language.

\subsection{Language Similarity}
\label{sec:lang_sim}
A intuitive hypothesis is that aspects of language similarity like the script or mutual intelligibility might affect the levels of instruction tuning cross-lingual transfer between languages. We test this using a case study of 7 Slavic languages, looking into possible effects of such aspects. However, we do not find a signal indicating these factors strongly correlate with cross-lingual transfer for this setting.

\paragraph{Setup} We train models on monolingual versions of the data in Russian, Serbian, Croatian, Slovenian, Polish, Slovak and Czech, and evaluate their transfer to each other. These languages can be divided along several linguistic lines that are summarized in Table~\ref{tab:lang_sim}. First, Russian is East Slavic, and the rest are either South or West Slavic. Second, Russian and Serbian both use the Cyrillic script, while the rest use Latin. Moreover, both Serbian and Croatian, and Slovak and Czech share a significant degree of mutual intelligibility.

\begin{table}[t]
    \small
    \centering
    \begin{tabular}{@{}lcccc@{}}
    \toprule
   \multirow{2}{*}{\textbf{Language}} & \multirow{2}{*}{\textbf{Code}} & \multicolumn{1}{@{}c}{\textbf{Slavic}} &\multirow{2}{*}{\textbf{Script}} &  \multicolumn{1}{@{}c}{\textbf{Mutually }}    \\
        & & \multicolumn{1}{@{}c}{\textbf{Family}} &  & \multicolumn{1}{@{}c}{\textbf{Intelligible}} \\
    \midrule
    Russian & ru & East & Cyrillic & -   \\
    Serbian & sr & South & Cyrillic & Croatian  \\
    Croatian & hr & South & Latin & Serbian   \\
    Slovenian & sl & South & Latin & -   \\
    Polish & pl	 & West & Latin &	-  \\
    Slovak & sk	& West & Latin & Czech  \\
    Czech &	cs	& West & Latin & Slovak  \\
    \bottomrule
    \end{tabular}
    \caption{Languages used for language similarity experiment, along with their language code, subfamily, script, and the language they are mutually intelligible with.}
    \label{tab:lang_sim}
\end{table}

\begin{figure}[t!]
\centering
  \includegraphics[width=0.99
  \linewidth]{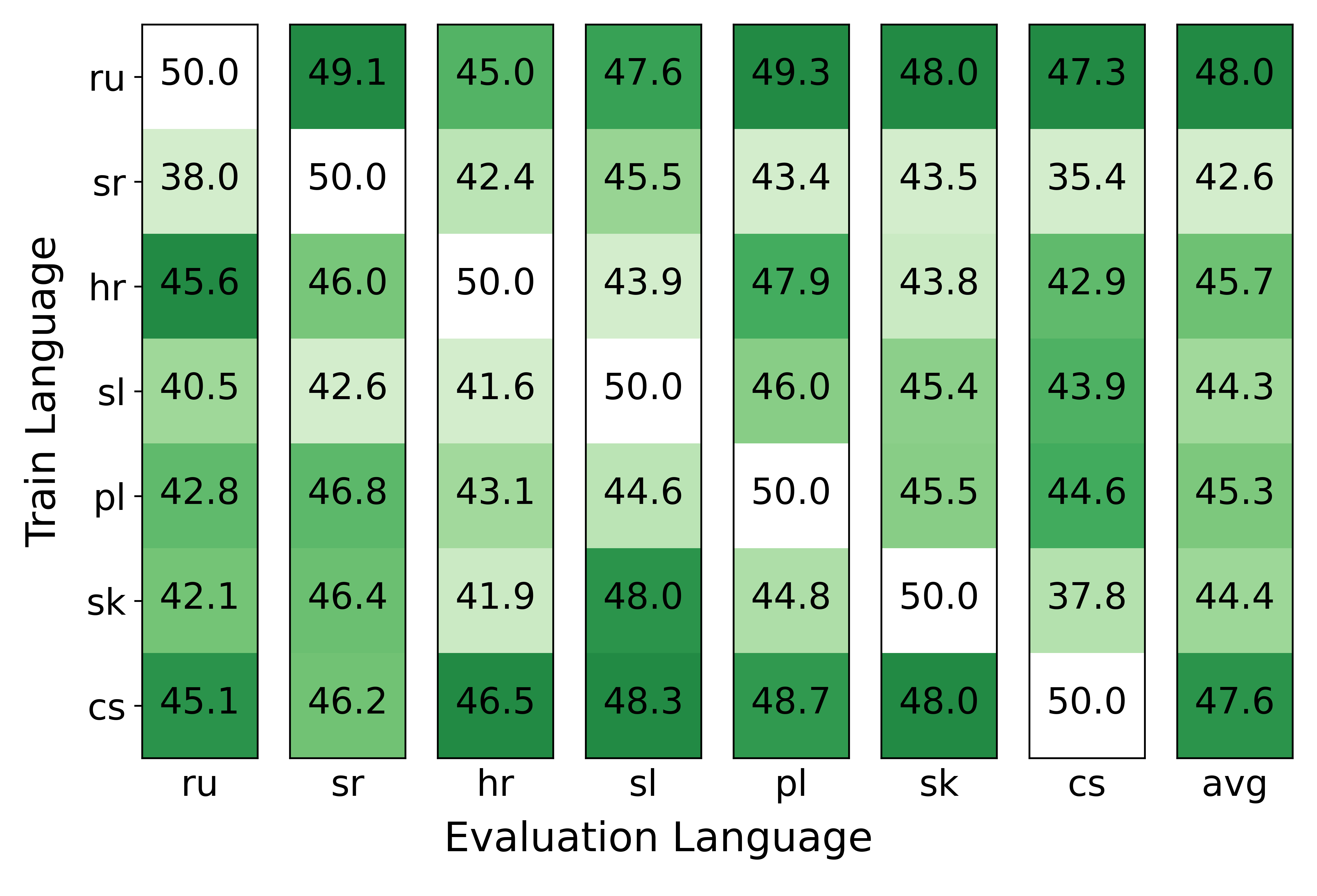}
  \caption{Instruction-following scores per language of models tuned monolingually. Each row represents a model trained using a different language, and each column is an individual heatmap of the scores of all models on the same evaluation language. The scores along the diagonal are excluded from the heatmaps coloring.}
\label{fig:langs_sim} 
\end{figure}
\paragraph{Results} Results are displayed on Figure~\ref{fig:langs_sim}. As shown, there is no a strong signal indicating that any of the aspects above is correlated with better mutual cross-lingual transfer. Russian and Czech tend to transfer instruction-following abilities best, and even though Russian and Serbian both use Cyrillic, Croatian and Czech transfer capabilities to Russian better than Serbian. Examining the effect of mutual intelligibility, Croatian and Serbian do not share cross-lingual abilities more than other languages, and while Slovak and Czech are mutually intelligible, Slovak transfers to Czech less than the rest. Our results align with recent findings that language similarity does not impact transferability or interference in machine translation given sufficient data and model capacity \cite{fernandes2023scaling,shaham-etal-2023-causes}.

\subsection{Fraction of Data in Pre-training}
\begin{figure}[t!]
\centering
   \includegraphics[width=0.99
   \linewidth]{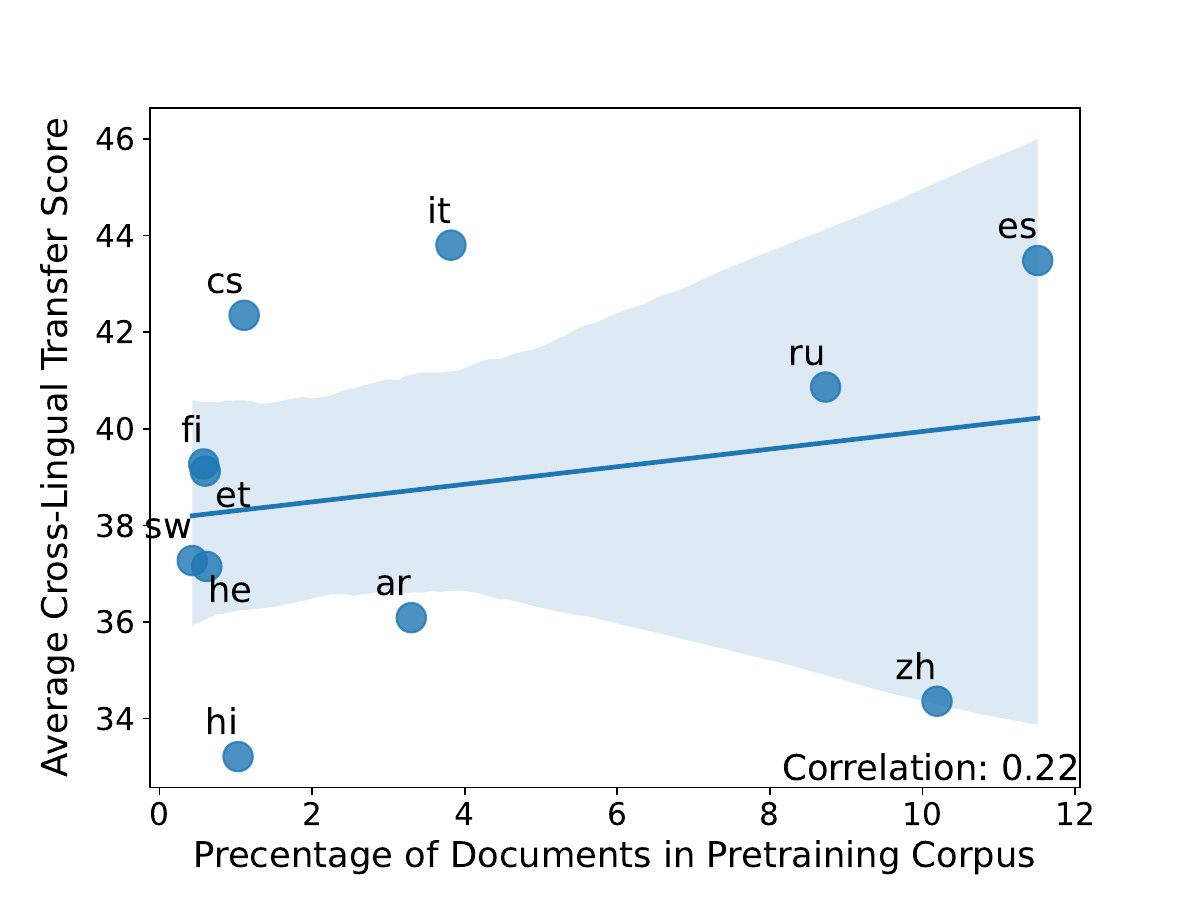}
  \caption{Weak Pearson correlation between the percentage of documents in the pre-training corpus (excluding English), and the average instruction-following score across languages for every training language. Blue area around the line is the confidence interval.}
\label{fig:corr} 
\end{figure}

A second possible predictor of the degree of cross-lingual transfer from a particular language is the extent to which the model was exposed to it during pre-training. Generally, a model's downstream performance on a specific language correlates with the fraction of data in that language in the pre-training corpus \citep{muennighoff-etal-2023-crosslingual}.  In contrast, Figure~\ref{fig:corr} suggests this is not necessarily the case for the cross-lingual transfer from a specific language. We find a weak Pearson correlation of 0.22 between the average cross-lingual score of each language and the number of documents in that language in pre-training corpus (Table 21 in \citet{anil2023palm}).

\section{Related work}
\paragraph{Cross-lingual Transfer}
The success of the pre-training--fine-tuning paradigm \cite{devlin-etal-2019-bert} ignited a new line of work on cross-lingual transfer. \citet{pires-etal-2019-multilingual} and \citet{wu-dredze-2019-beto} showed that the multilingual variant of BERT can be fine-tuned on a specific task in one language and preform this task on another language, and \citet{artetxe-schwenk-2019-massively} reported similar findings with a Recurrent Neural Network. \citet{conneau-etal-2020-unsupervised} introduced XLM-R, a multilingual pre-trained encoder with strong cross-lingual abilities. \citet{phang-etal-2020-english} showed that intermediate training on an English task improves XLM-R's transfer across languages further, and \citet{pfeiffer-etal-2020-mad} suggested an adapter-based framework to improve cross-lingual and task generalization. \citet{pmlr-v119-hu20b} proposed a benchmark for cross-lingual generalization consists of 40 languages across 9 NLP tasks. 

\citet{K2020Cross-Lingual} found that the depth of the network matters for cross-lingual transfer, and \citet{conneau-etal-2020-emerging} showed that parameter sharing is more important than shared vocabulary.  \citet{choenni-etal-2023-languages} delved into the influence of specific examples from the training data on the performance in other languages, and \citet{malkin-etal-2022-balanced} investigated how pre-training BERT-based models using different language pairs affects cross-lingual downstream performance. Going beyond encoder-only models, \citet{xue-etal-2021-mt5} proposed mT5, a multilingual variant of T5 \cite{JMLR:v21:20-074}, and showed the significance of model scaling for cross-lingual transfer in generation tasks. \citet{ye2023language} explored trasferability in English-centric models \cite{touvron2023llama} using four tasks.

In contrast to most cross-lingual transfer literature that is focused on task-specific fine-tuning, we explore trends of cross-lingual generalization for general-purpose instruction-following LLMs.

\paragraph{Multilingual Instruction Tuning}
Initially, works on instruction tuning \cite{mishra-etal-2022-cross,wei2022finetuned,sanh2022multitask} focused on cross-task generalization in English. Subsequently, a large body of work was dedicated to multilingual instruction tuning. \citet{muennighoff-etal-2023-crosslingual} found that tuning models with English datasets enables zero-shot cross-lingual abilities to new languages. The authors also found that this holds for languages that the model has never intentionally seen during pre-training, and that multilingual training improves generalization to new tasks. \citet{chen2023monolingual} investigated the effects of full parameter training vs low-rank adaptation \cite{hu2022lora} and monolingual vs multilingual instruction tuning using the Stanford Alpaca \cite{alpaca} data, machine translated into 5 languages. \citet{lai-etal-2023-okapi} trained multilingual instruction-following models for 26 languages with reinforcement learning from human feedback \cite{NEURIPS2022_b1efde53}, and \citet{zhang2023plug} suggested instruction tuning LLMs by prepending the instruction and response translated into a pivot language (e.g English) to the response in the target language. Concurrently with our work, \citet{kew2023turning} found that only a few languages in the tuning set result in better cross-lingual transfer to new languages for English-centric LLMs.

In this work, we consider transfer from monolingual instruction tuning from 12 languages, rather than exclusively on English. Furthermore, we examine multilingual instruction-following using an LLM pre-trained on hundreds of languages, which might be a key to  unlocking more transfer to languages not represented during tuning. Importantly, we unveil the potential of just a small amount of language diversity in the instruction tuning set for this cross-lingual generalization.

\section{Conclusion}
We demonstrate that cross-lingual transfer offers a promising avenue for building multilingual instruction-following LLMs. Our findings across different languages suggest that even monolingual instruction tuning using only one language can result in improved instruction-following capabilities in other languages. Moreover, incorporating even a small set of a few dozen multilingual examples can significantly enhance instruction-following performance for both the languages the model is tuned on, and ones that were only seen during pre-training. Additionally, training on such multilingual datasets achieves comparable or even superior performance compared to monolingual tuning for some languages. We observe a similar trend when exploring the effect of total number of languages in the tuning set, as even splitting the train set to only two languages improves generalization to new languages, compared to monolingual tuning. These findings pave the way for efficient and scalable development of multilingual LLMs capable of understanding and following instructions across languages with minimal multilingual supervision.

\section{Limitations}
Limitations of our work include the use of translation for expanding datasets to multilingual settings, the number of languages we evaluated on, and number of models we experimented with. We now discuss each of them.

\paragraph{Translated data} One limitation of our work is that our data is translated using the Google Translate API, and not originally sourced by native speakers. Automatic translation is inherently imperfect and may introduce noise to the tuning sets. However, translation also allows to for a controlled setup with parallel data, in which the content of all training and evaluation examples is the same for all languages.

\paragraph{Number of languages} A second limitation is that we use 12 languages in our main experiments (\S\ref{sec:how_much}), with 3 additional languages in the language similarity experiment (\S\ref{sec:lang_sim}). Clearly, multilingual instruction-following models need to successfully operate in many more languages, and we leave work on scaling this number to future work.

\paragraph{Number of models} Lastly, we experiment with PaLM 2, and results may vary with different LLMs. Nevertheless, our focus on PaLM 2 highlights the potential of multilingual pre-training for future advancements in LLMs.

\section*{Acknowledgments}
We thank Omer Levy, Or Honovich, Alon Jacovi, Avi Caciularu, and Omer Goldman for their valuable feedback.

\bibliographystyle{acl_natbib}
\bibliography{custom}

\appendix

\section{Languages}
\label{sec:langs}
The languages we use, their language families, scripts ,and language codes are shown in Table~\ref{tab:lang_codes}.

\section{Side-By-Side Evaluation}
\label{sec:sbs_prompt}
Figure~\ref{fig:prompt} shows the prompt given the the LLM judge for the side-by-side evaluation.

\begin{table}[t]
  \small
  \centering
  \begin{tabular}{@{}llll@{}}
  \toprule
  \textbf{Language} & \textbf{Code} & \textbf{Family} & \textbf{Script} \\
  \midrule
  Arabic & ar & Afro-Asiatic & Arabic \\
  Chinese & zh & Sino-Tibetan & Chinese \\
  Czech & cs & Indo-European & Latin \\
  English & en & Indo-European & Latin \\
  Estonian & et & Uralic & Latin \\
  Finnish & fi & Uralic & Latin \\
  Hebrew & he & Afro-Asiatic & Hebrew \\
  Hindi & hi & Indo-European & Devanagari \\
  Italian & it & Indo-European & Latin \\
  Russian & ru & Indo-European & Cyrillic \\
  Spanish & es & Indo-European & Latin \\
  Swahili & sw & Niger-Congo & Latin \\
  \bottomrule
  \end{tabular}
  \caption{Languages used in our main experiments.}
  \label{tab:lang_codes}
\end{table}

\section{Training and Inference Details}
\label{sec:hyperparams}
We now describe the hyperparameters we use in our experiments.
We tune every model for 2,000 steps, using a fixed learning rate of 1e-5, a batch size of 128, and a dropout rate of 0.05. We limit inputs to 1,024 tokens and targets to 512 tokens. We sample a development set of 250 examples from every training set and select the checkpoint based on the development RougeL \cite{lin-2004-rouge} score. During inference, we generate responses of up to 512 tokens using nucleus sampling \citep{Holtzman2020The} with $p=0.9$ and temperature of 0.7. For the judge, we use greedy decoding to generate the ID of the better response (1 or 2).

\begin{figure}
\small
\centering
{\setlength{\extrarowheight}{3pt}%
\begin{tabular}{|p{0.94\columnwidth}|} 
\hline
\textcolor[HTML]{000000}{Below is an instruction and two answers. Choose your preferred answer, which can be subjective.}\\\\

\textcolor[HTML]{000000}{The instruction:}\\
\textcolor[HTML]{000000}\{instruction\}\\\\

\textcolor[HTML]{000000}{Answer1:}\\
\textcolor[HTML]{000000}\{response 1\}\\\\

\textcolor[HTML]{000000}{Answer2:}\\
\textcolor[HTML]{000000}\{response 2\}\\\\

\textcolor[HTML]{000000}{Which one is better, Answer1 or Answer2?}\\
\textcolor[HTML]{000000}{Only write a single digit as your answer, '1' for Answer1 or '2' for Answer2. Do not add any explanation.}
\vspace{2pt}
\\
\hline
\end{tabular}}
\caption{Side-by-side evaluation prompt.}
\label{fig:prompt}
\end{figure}

\section{Judge-Human Agreement}
\label{sec:agreement}
To measure PaLM~2-L agreement with human judgments across language, we conduct a human annotation process on four languages, English, Spanish, Russian, and Hebrew. For every language we sample 50 instructions and let two native speakers select the better response out of two options, similarly to the task we assign the LLM judge (Figure~\ref{fig:prompt}). We always present the response by the model that was monolingually tuned using the evaluation language, alongside a response by model selected at random from the of the monolingually tuned ones described in~\S\ref{sec:mono}. The agreement score on a single instruction is 1 if the LLM judge and human agree, 0.5 if exactly one of them selects a tie, and 0 if each selects a different response \citep{zhou2023lima}. Table~\ref{tab:agreement} shows the results. Overall, the LLM judge agreement with humans is strong for all four languages, yet there is some room of 2.5-7 points from inter human agreement in all languages. As expected, the models' highest agreement with humans is in English with 79.5\%,. In the rest of the languages the agreement is a few points lower.

\begin{table}[t]
    \small
    \centering
    \begin{tabular}{@{}ccc@{}}
    \toprule
\textbf{Language} &  \textbf{Human-Model} & \textbf{Human-Human} \\ 
  \midrule
  English  & 79.5 & 85.0 \\
  Spanish & 77.0 & 80.0 \\
  Russian  & 76.5 & 79.0  \\
  Hebrew & 75.0 & 82.0 \\
  \bottomrule
  \end{tabular}
  \caption{Judges agreement scores per language.}
  \label{tab:agreement}
\end{table}

\section{Response Language}
\label{sec:response_lang}
When a user prompts a model in a specific language, they usually expect to receive a response in that same language. However, pre-trained LLMs often respond in a different language than the language of their prompt \cite{touvron2023llama,chen2023monolingual,kew2023turning}. This poses a challenge also for evaluation of open-ended queries, since those are commonly evaluated with an LLM-as-a-judge \cite{zheng2023judging} protocol, and the judges often ignore whether the response language match the prompt language, even when instructed not to \cite{chen2023monolingual}. Usually, this is handled by forcing the lowest score to such response \cite{chen2023monolingual,kew2023turning}, which does not account for all cases.\footnote{For example, a response in English to a prompt in French can still be very helpful, or when the prompt is a request for translation or code.} To verify our trained models respond in the same language as their prompt, we manually annotate the language of responses to evaluation instructions in all languages. For every language, we randomly sample 20 responses from the pool of models tuned monolingually in other languages, to end up with a total of 240 generations from various models. We find that 239 responses are in the same language as the prompt, as desired. This is a major difference in the behavior of our PaLM~2-based instruction-tuned models and the commonly used  \cite{chen2023monolingual,kew2023turning} LLaMA-based ones \cite{touvron2023llama,touvron2023llama2}. We hypothesize this stems from the multilingual emphasis in the pre-training of PaLM~2, compared to the more English-centric LLaMA.

\begin{figure*}[t!]
\centering
  \includegraphics[width=0.99
  \linewidth]{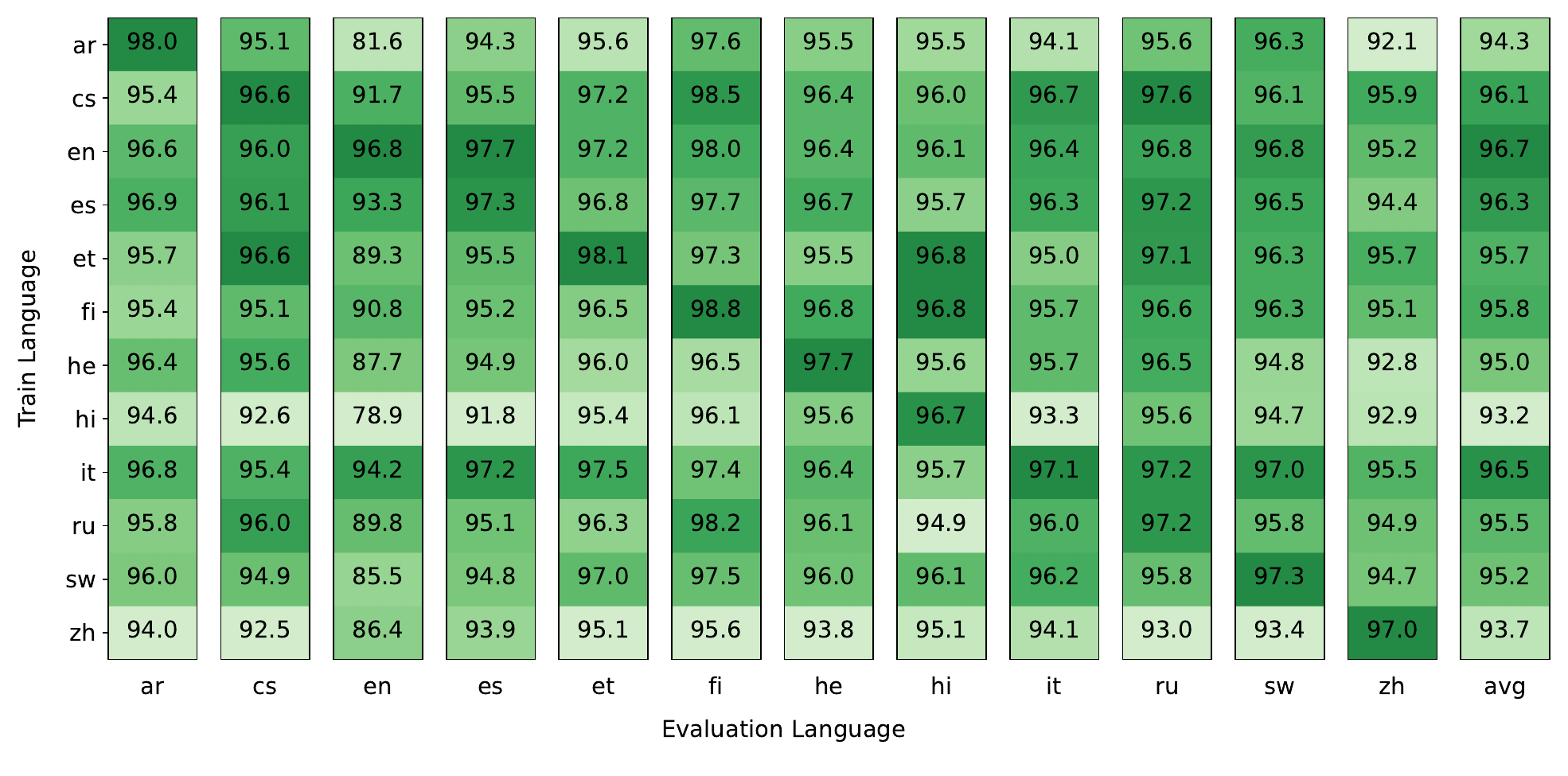}
  \caption{Per language instruction-following comparisons of models instruction-tuned on monolingual data to the pre-trained model that was not instruction-tuned. Each row
represents a model tuned using a different language, and each column is an individual heatmap of the scores of
all models on the same evaluation language. Scores are the discounted-ties weighted average of the side-by-side
scores against the pre-trained model.}
\label{fig:heatmaps_base} 
\end{figure*}

\section{Comparison to The Base Model}
\label{sec:base_model}
The scores of models of model instruction-tuned monolingually compared to the pre-trained model that was not instruction-tuned, as opposed to our main evaluation setup, are shown in Figure~\ref{fig:heatmaps_base}. As evident, instruction tuning the model on each of the languages separately unlocks instruction-following abilities across all languages.

\section{Languages Permutations}
\label{sec:lang_perms}
 We use 3 different permutations of 6 languages to determine the order in which we add languages to the tuning set in the experiment described Section~\ref{sec:add_langs}. The permutations are displayed in Table~\ref{tab:add_langs}.

\begin{table}[t]
    \small
    \centering
    \begin{tabular}{@{}cccccc@{}}
    \toprule
    \textbf{1} & \textbf{2} & \textbf{3} & \textbf{4} & \textbf{5} & \textbf{6} \\
    \midrule
    fi & fi,en & fi,en,ru & fi,en,ru,it &   fi,en,ru,it,sw & all six \\
    sw & sw,it &  sw,it,ar & sw,it,ar,en &  sw,it,ar,en,fi & all six\\
    it & it,fi & it,fi,en & it,fi,en,ar & it,fi,en,ar,ru  &  all six\\
    \bottomrule
    \end{tabular}
    \caption{Subsets of languages used to tune models for the experiment described in Section~\ref{sec:add_langs}. Each cell represents a version of the training set, for which all examples are uniformly split between the languages in that cell.}
    \label{tab:add_langs}
\end{table}

\end{document}